%% file: naaclhlt2019.tex
\documentclass[11pt,a4paper]{article}
\usepackage[nohyperref]{naaclhlt2019}
\usepackage{times}
\usepackage{latexsym}
\usepackage{url}
\usepackage{multirow}
\usepackage{amsmath,amssymb}
\usepackage{ifthen}
\usepackage{graphicx}
\usepackage{stmaryrd}
\usepackage{enumitem}
\usepackage{multicol}
\usepackage{color}
\usepackage{subfigure}
\usepackage{algorithm}
\usepackage[noend]{algpseudocode}
\usepackage[font={footnotesize}]{caption}

\input std-macros

\input macros

\newcommand\ourmodel{\textsc{VBSiX}}
\newcommand\ourmodels{\textsc{VBSiX }}

\newcommand\scenedomain{\textsc{Scene}}
\newcommand\tangramdomain{\textsc{Tangram}}
\newcommand\alchemydomain{\textsc{Alchemy}}
\newcommand\mml{\textsc{MML}}
\newcommand\reinforce{\textsc{REINFORCE}}
\newcommand\expertmml{\textsc{Expert-MML}}

\aclfinalcopy % Uncomment this line for the final submission
\title{Value-based Search in Execution Space for Mapping Instructions to Programs}

\author{Dor Muhlgay$^{1}$ ~~~~~ Jonathan Herzig$^{1}$ ~~~~~
Jonathan Berant$^{1,2}$ \\
\mbox{}\\
$^1$School of Computer Science, Tel-Aviv University \\
$^2$Allen Institute for Artificial Intelligence \\
\small{\texttt{\{dormuhlg@mail,jonathan.herzig@cs,joberant@cs\}.tau.ac.il}}}

\date{}

\begin{document}
\maketitle

\input{00_abstract}
\input{01_intro}

\input{02_task}

\input{03_mdp}

\input{04_exbs}

\input{05_training}

\input{06_experiments}

\input{07_related}

\input{08_conclusions}

\bibliography{all.bib}
\bibliographystyle{acl_natbib.bst}

\appendix
\input{09_nn.tex}

\input{10_hyper.tex}

\input{11_prev.tex}
\input{12_path.tex}

\end{document}

%% file: std-macros.tex
\newcommand\sA{\ensuremath{\mathcal{A}}}

\newcommand\sD{\ensuremath{\mathcal{D}}}
\newcommand\sE{\ensuremath{\mathcal{E}}}

\newcommand\sO{\ensuremath{\mathcal{O}}}

\newcommand\sS{\ensuremath{\mathcal{S}}}
\newcommand\sT{\ensuremath{\mathcal{T}}}

\newcommand\sV{\ensuremath{\mathcal{V}}}

\newcommand\sZ{\ensuremath{\mathcal{Z}}}

\newcommand\bh{\ensuremath{\mathbf{h}}}

\newcommand\bu{\ensuremath{\mathbf{u}}}

\newcommand\bz{\ensuremath{\mathbf{z}}}

\newcommand\refsec[1]{Section~\ref{sec:#1}}

\newcommand\reffig[1]{Figure~\ref{fig:#1}}

\newcommand\reftab[1]{Table~\ref{tab:#1}}

\newcommand\refalg[1]{Algorithm~\ref{alg:#1}}

\ifthenelse{\isundefined{\definition}}{\newtheorem{definition}{Definition}}{}
\ifthenelse{\isundefined{\assumption}}{\newtheorem{assumption}{Assumption}}{}
\ifthenelse{\isundefined{\proposition}}{\newtheorem{proposition}{Proposition}}{}
\ifthenelse{\isundefined{\theorem}}{\newtheorem{theorem}{Theorem}}{}
\ifthenelse{\isundefined{\lemma}}{\newtheorem{lemma}{Lemma}}{}
\ifthenelse{\isundefined{\corollary}}{\newtheorem{corollary}{Corollary}}{}
\ifthenelse{\isundefined{\alg}}{\newtheorem{alg}{Algorithm}}{}
\ifthenelse{\isundefined{\example}}{\newtheorem{example}{Example}}{}
\DeclareMathOperator*{\E}{\mathbb{E}} % Diagnonal matrix

%% file: macros.tex
\newcommand\den[2]{{\llbracket #1 \rrbracket}_{#2}} % Denotation
\newcommand\denp[2]{{\llbracket #1 \rrbracket}^{ex}_{#2}} % Partial denotation

%% file: 00_abstract.tex
\begin{abstract}
Training models to map natural language instructions to programs, given target world supervision only, requires searching for good programs at training time. Search is commonly done using beam search in the space of partial programs or program trees, but as the length of the instructions grows finding a good program becomes difficult. In this work, we propose a search algorithm that uses the target world state, known at training time, to train a \emph{critic} network that predicts the expected reward of every search state. We then score search states on the beam by interpolating their expected reward with the likelihood of programs represented by the search state. Moreover, we search not in the space of programs but in a more compressed state of program executions, augmented with recent entities and actions. On the SCONE dataset, we show that our algorithm dramatically improves performance on all three domains compared to standard beam search and other baselines.
\end{abstract}

%% file: 01_intro.tex
\section{Introduction} 
Training models that can understand natural language instructions and execute them in a real-world environment is of paramount importance for communicating with virtual assistants and robots, and therefore has attracted considerable attention \cite{branavan09reinforcement,vogel10navigate,chen11navigate}.
A prominent approach is to cast the problem as semantic parsing, where instructions are mapped to a high-level programming language \cite{artzi2013weakly,long2016projections,guu2017bridging}. Because annotating programs at scale is impractical, it is desirable to train a model from instructions, an initial world state, and a target world state only, letting the program itself be a latent variable.

Learning from such weak supervision results in a difficult search problem at training time. The model must search for a program that when executed leads to the correct target state. Early work employed lexicons and grammars to constrain the search space \cite{clarke10world,liang11dcs,krishnamurthy2012weakly,berant2013freebase,artzi2013weakly}, but recent success of sequence-to-sequence models \cite{sutskever2014sequence} shifted most of the burden to learning. Search is often performed simply using beam search, where program tokens are emitted from left-to-right, or program trees are generated top-down \cite{krishnamurthy2017neural,yin2017syntactic,cheng2017learning,rabinovich2017abstract} or bottom-up \cite{liang2017nsm,guu2017bridging,goldman2018weakly}. Nevertheless, when instructions are long and complex and reward is sparse, the model may never find enough correct programs, and training will fail.

In this paper, we propose a novel search algorithm for mapping a sequence of natural language instructions to a program, which extends the standard beam-search in two ways. First, we capitalize on the target world state being available at training time and train a \emph{critic} network that given the language instructions, current world state, and target world state estimates the expected future reward for each search state. In contrast to traditional beam search where states representing partial programs are scored based on their likelihood only, we also consider expected future reward, leading to a more targeted search at training time. Second, rather than search in the space of programs, we search in a more compressed execution space, where each state is defined by a partial program's execution result. %while tracking recent entities and actions. This makes the search space smaller, while allowing us to handle references to past entities and actions.

We evaluated our method on the SCONE dataset, which includes three different domains where long sequences of 5 instructions are mapped to programs. We show that while standard beam search gets stuck in local optima and is unable to discover good programs for many examples, our model is able to bootstrap, improving final performance by 20 points on average. We also perform extensive analysis and show that both value-based search as well as searching in execution space contribute to the final performance. Our code and data are
available at \url{http://gitlab.com/tau-nlp/vbsix-lang2program}.

%% file: 02_task.tex
\section{Background}
\label{sec:background}

% Define context
Mapping instructions to programs invariably involves a \emph{context}, such as a database or a robotic environment,
in which the program (or logical form) is executed.
% Training goal
The goal is to train a model given a training set 
$\{ (x_{(j)} = (c_{(j)}, \bu_{(j)}), y_{(j)}) \}_{j=1}^N$,
where $c$ is the context, $\bu$ is a sequence of natural language instructions, and $y$ is the target state of the environment after following the instructions, which we refer to as \emph{denotation}.
The model is trained to map the instructions $\bu$ to a program $\bz$ such that executing $\bz$ in the context $c$ results in the denotation $y$, which we denote by $\den{\bz}{c} = y$. 
Thus, the program $\bz$ is a latent variable we must search for at both training and test time. When the sequence of instructions is long, search becomes hard, particularly in the early stages of training.
%The task is to train a model that receives the context $c$ and a natural language instructions $u$, and maps $u$ to a program $z$ such that applying $z$ with $c$ will result in a target result $y$. Each training example labels each utterance and context with its target result: $\{ (x_{(j)} = (c_{(j)}, u_{(j)}), y_{(j)}) \}_{j=1}^N$. Thus, the program $z$ is a latent variable we must search for at both training time and test time.

Recent work tackled the training problem using variants of reinforcement learning (RL) \cite{Suhr2018Situated,liang2018mapo} or maximum marginal likelihood (MML) \cite{guu2017bridging,goldman2018weakly}.
We now briefly describe MML training, which we base our training procedure on, and outperformed RL in past work under comparable conditions \cite{guu2017bridging}.

%Standard approach for tackling those tasks is by using maximum marginal likelihood (MML) method to train a generative model. 
We denote by $\pi_{\theta}(\cdot)$ a model, parameterized by $\theta$, that generates the program $\bz$ token by token from left to right. The model $\pi_{\theta}$ receives the context $c$, instructions $\bu$ and previously predicted tokens $z_{1 \ldots t-1}$, and returns a distribution over the next program token $z_t$.  
The probability of a program prefix is defined to be:  $p_{\theta}(z_{1 \ldots t} \mid \bu, c) = \prod_{i=1}^t \pi_{\theta}(z_{i} \mid \bu, c, z_{1 \ldots i-1})$.  The model is trained to maximize the MML objective: 
\begin{equation*}\label{eq:pareto mle2}
\begin{aligned}
J_{MML} = \log \prod_{(c, \bu, y)} p_\theta(y \mid c, \bu) =  \\
 \sum_{(c, \bu, y)} \log (\sum_\bz p_\theta(\bz \mid \bu,c)\cdot R(\bz)),
 \end{aligned}
\end{equation*}
where $R(\bz) = 1$ if $\den{\bz}{c} = y$, and 0 otherwise (For brevity, we omit $c$ and $y$ from $R(\cdot)$). Typically, the \mml{} objective is optimized with stochastic gradient ascent, where the gradient for an example $(c, \bu, y)$ is: 

$$\nabla_{\theta} J_{MML} = \sum_{z} q(z)\cdot R(z) \nabla_{\theta} \log p_\theta(z|c, \bu)$$
$$q(z) := \frac{R(z)\cdot p_\theta(z|c, \bu)}{\sum_{\tilde{z}}R(\tilde{z})\cdot p_\theta(\tilde{z}|c, \bu)}. $$
%\propto p_{\theta}(z|c, \bu)$$

The search problem arises because it is impossible to enumerate the set of all programs, and thus the sum over programs is approximated by a small set of high probability programs, which have high weights $q(\cdot)$ that dominate the gradient.
Search is commonly done with \textit{beam-search}, an iterative algorithm that builds an approximation of the highest probability programs according to the model. At each time step $t$, the algorithm constructs a beam $B_t$ of at most $K$ program prefixes of length $t$. Given a beam $B_{t-1}$, $B_t$ is constructed by selecting the $K$ most likely \emph{continuations} of prefixes in $B_{t-1}$ , according to $p_{\theta}(z_{1..t} | \cdot)$. The algorithm runs $L$ iterations, and returns all complete programs discovered.

In this paper, our focus is the search problem that arises at training time when training from denotations, i.e., finding programs that execute to the right denotation.
Thus, we would like to focus on scenarios where programs are long, and standard beam search fails. We next describe the SCONE dataset, which provides such an opportunity.

%Our goal is to develop methods for mapping natural language to programs from goal state supervision only, where finding the latent program poses a difficult search challenge at training time. 
%Thus, we would like to focus on scenarios where the programs are long, leading to a difficult search problem. In such settings, the standard beam-search will not suffice to train the model. 

\begin{figure}
    \centering
    \includegraphics[width=8.3cm, height=4.3cm]{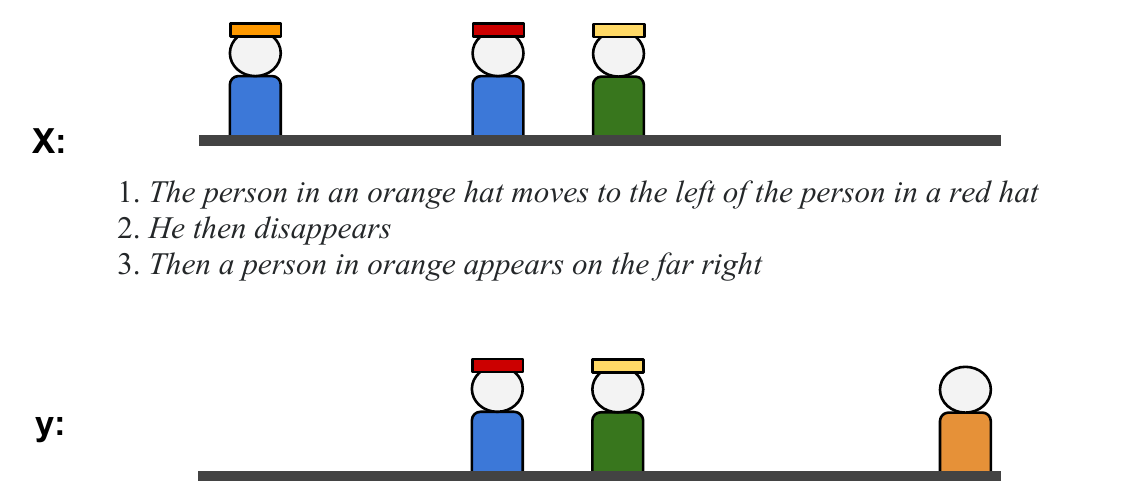}
    \caption{Example from the \scenedomain{} domain in SCONE (3 utterances), where people with different hats and shirts are located in different positions. Each example contains an initial world (top), a sequence of natural language instructions, and a target world (bottom). 
    %Intermediate worlds and commands are latent.
    }
    \label{fig:scone_example}
\end{figure}

%\FigTop{scone_example_short.pdf}{0.55}{scone_example}{
%Example from the SCENE domain in SCONE (3 utterances), where people with different hats and shirts are located in different positions. Each example contains an initial world (top), a sequence of natural language instructions, and a target world (bottom). Intermediate world states and programs instruction are latent.}

\begin{figure*}[t]
    \centering
    \includegraphics[width=11cm, height=4.5cm]{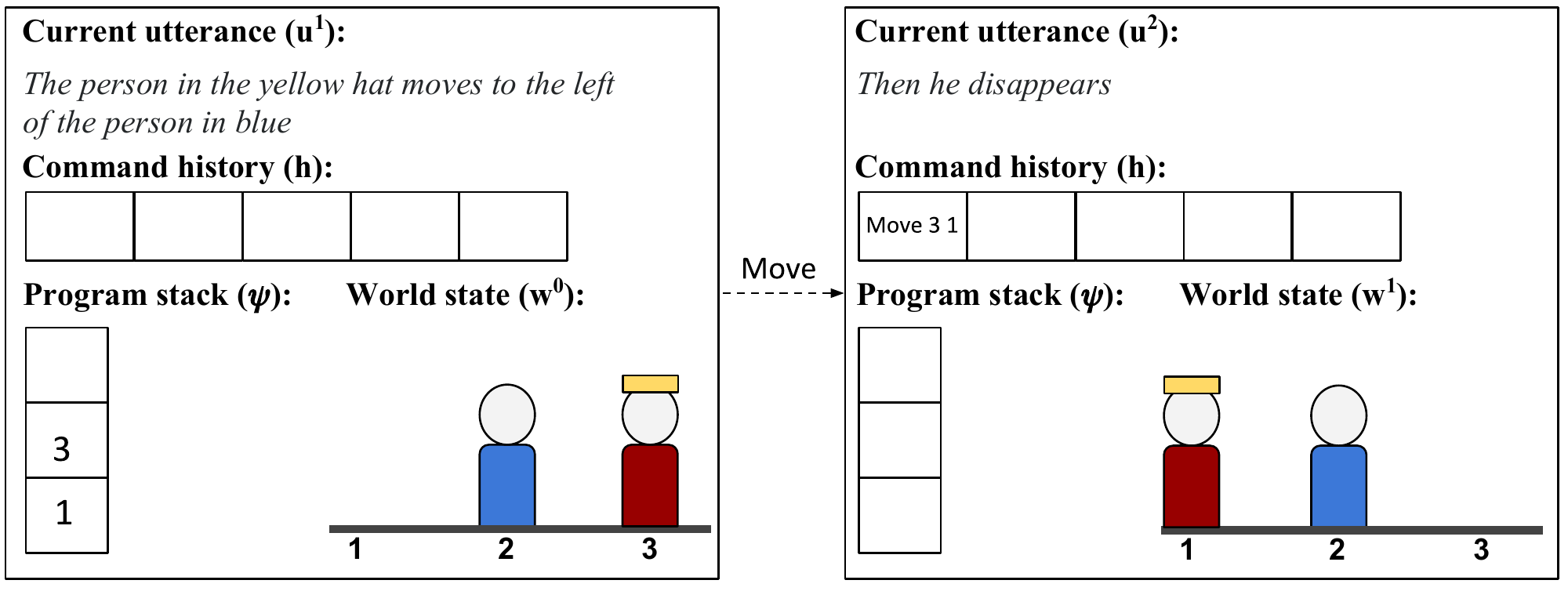}
    \caption{An illustration of a state transition in SCONE. $\pi_\theta$ predicted the action token \textsc{move}. Before the transition (left) the command history is empty, and the program stack contains the arguments 3 and 1 which were computed in previous steps. Note that the state does not include the partial program that computed those arguments. After transition (right) the executor popped the arguments from the program stack and applied the action \textsc{move 3 1}: the man with the yellow hat moved to position 1 and the action is added to the execution history with its arguments. Since this action terminated the command, $\pi_\theta$ advanced to the next utterance.}
    \label{fig:scone_state}
\end{figure*}

\paragraph{The SCONE dataset} 
\newcite{long2016projections} presented the SCONE dataset, where a  sequence of instructions needs to be mapped to a program consisting of a sequence of executable commands. The dataset has three domains, where each domain includes several objects (such as \emph{people} or \emph{beakers}), each with different properties (such as \emph{shirt color} or \emph{chemical color}). 
%Given 5 sequential instructions in natural language, the goal is to predict the final world state. 
SCONE provides a good environment for stress-testing search algorithms because a long sequence of instructions needs to be mapped to a program.
\reffig{scone_example} shows an example from the \scenedomain{} domain. 

Formally, the context in SCONE is a \emph{world} specified by a list of positions, where each position may contain an object with certain properties. A formal language is defined to interact with the world. The formal language contains \emph{constants} (e.g., numbers and colors), \emph{functions} that allow to query the world and refer to objects and intermediate computations, and \emph{actions}, which are functions that modify the world state. Each command is composed of a single action and several arguments constructed recursively from constants and functions. E.g., the command \textsc{move(hasHat(yellow), leftOf(hasShirt(blue)))}, contains the action \textsc{move}, which moves a person to a specified position. The person is computed by  \textsc{hasHat(yellow)}, which queries the world for the position of the person with a yellow hat, and the target position is computed by \textsc{leftOf(hasShirt(blue))}. We refer to \newcite{guu2017bridging} for a full description of the language.

Our goal is to train a model that given an initial world $w^0$ and a sequence of natural language \emph{utterances} $\bu = (u^1, \dots, u^M)$, will map each utterance $u^i$ to a command  $z^i$ such that applying the program $\bz = (z^1, \dots, z^M)$ on $w^0$ will result in the target world, i.e., $\den{\bz}{w^0} = w^M = y$.

%% file: 03_mdp.tex
\section{Markov Decision Process Formulation}\label{sec:mdp}
%To present our search algorithm, we first formulate the problem as a Markov Decision Process (MDP). %We show in \refsec{execution beam search} that the graph induced by an MDP when searching in the space of programs is a tree. Conversely, our algorithm will search in a more compact execution space, where the MDP corresponds to a graph.
To present our algorithm, we first formulate the problem as a Markov Decision Process (MDP), which is a tuple $(\sS, \sA, R, \delta)$. %where $\sS$ is the state space, $\sA$ is the action space, $R(\cdot)$
%is the reward function, and $\delta(\cdot)$ is a deterministic transition function. 
To define the state set $\sS$,  we assume all program prefixes are executable, which can be easily done as we show below.
The \emph{execution result} of a prefix $\tilde{z}$ in the context $c$, denoted by $\denp{\tilde{z}}{c}$, contains its denotation and additional information stored in the executor. Let $\sZ_{\text{pre}}$ be the set of all valid programs prefixes. The set of states is defined to be $\sS = \{(x, \denp{\tilde{z}}{c}) \mid \tilde{z}\in \sZ_\text{pre} \}$, i.e., 
the input paired with all possible execution results. %Because many program prefixes have the same execution result, the \emph{execution space} $\sS$ is much smaller than the space of programs.

The action set $\sA$ includes all possible program tokens
\footnote{Decoding is constrained to valid program tokens only.}, and the transition function $\delta:  \sS \times \sA \rightarrow \sS$ is computed by the executor. Last, the reward $R(s,a) = 1$ iff the action $a$ ends the program and leads to a state $\delta(s,a)$ where the denotation is equal to the target $y$.
The model $\pi_{\theta}(\cdot)$ is a parameterized policy that provides a distribution over the program vocabulary at each time step. 

\paragraph{SCONE as an MDP} We define the partial execution result $\denp{\tilde{z}}{c}$ in SCONE, as described by \newcite{guu2017bridging}. We assume that SCONE's formal language is written in postfix notations, e.g., the instruction \textsc{move(hasHat(yellow), leftOf(hasShirt(blue)))} is written as \textsc{yellow hasHat blue hasShirt leftOf move}. With this notation, a partial program can be executed left-to-right by maintaining a \emph{program stack}, $\psi$. The executor pushes constants (\textsc{yellow}) to $\psi$, and applies functions (\textsc{hasHat}) by popping their arguments from $\psi$ and pushing back the computed result. Actions (\textsc{move}) are applied by popping arguments from $\psi$ and performing the action in the current world. 

To handle references to previously executed commands, SCONE's formal language includes functions that provide access to actions and arguments in previous commands. To this end, the executor maintains an \emph{execution history}, $\bh^i = (e^1, \ldots, e^{i})$, a list of executed actions and their arguments. Thus, the execution result of a program prefix is $\denp{\tilde{z}}{w^0} = (w^{i-1}, \psi, \bh^{i-1})$, which includes the current world, the program stack and the execution history. 

We adopt the model from  \newcite{guu2017bridging}  (architecture details in appendix~\ref{sec:nn}): The policy $\pi_\theta$ observes $\psi$ and $u^i$, the current utterance being parsed, and predicts a token. When the model predicts an action token that terminates a command, the model moves to the next utterance (until all utterances have been processed). The model uses functions to query the world $w^{i-1}$ and history $\bh^{i-1}$. %(querying $\bh^{i-1}$ allows the model to handle conference). 
Thus, each MDP state in SCONE is a pair $s=(u^i,\denp{\tilde{z}}{w^0})$. \reffig{scone_state} illustrates a state transition in the \scenedomain{} domain. Importantly, the state does not store the full program's prefix, and many different prefixes may lead to the same state. Next, we describe a search algorithm for this MDP.

%% file: 04_exbs.tex
\section{Searching in Execution Space} \label{sec:execution beam search}

%\FigTop{search_space.pdf}{0.4}{search_space}{
%A set of commands represented in program space (top) and execution space (bottom). Since multiple prefixes have the same execution result (e.g., \textsc{yellow hasHat} and \textsc{red hasShirt}), the execution space smaller. 
%}

%The represented instructions are: (1) \textsc{yellow hasHat blue hasShirt leftOf move}, (2) \textsc{yellow hasHat 1  move}, (3) \textsc{red hasShirt blue hasShirt leftOf move} and (4) \textsc{red hasShirt 1 move}. The instructions' context is the world state shown of the first state in \reffig{scone_state}%

Model improvement relies on generating correct programs given a possibly weak model. Standard beam-search explores the space of all program token sequences up to some fixed length. We propose two technical contributions to improve search: (a) We simplify the search problem by searching for correct executions rather than correct programs; (b) We use the target denotation at training time to better estimate partial program scores in the search space.  We describe those next.        

\subsection{Reducing program search to execution search}

\begin{figure}[t]
    \centering
    \includegraphics[width=8.3cm, height=4.7cm]{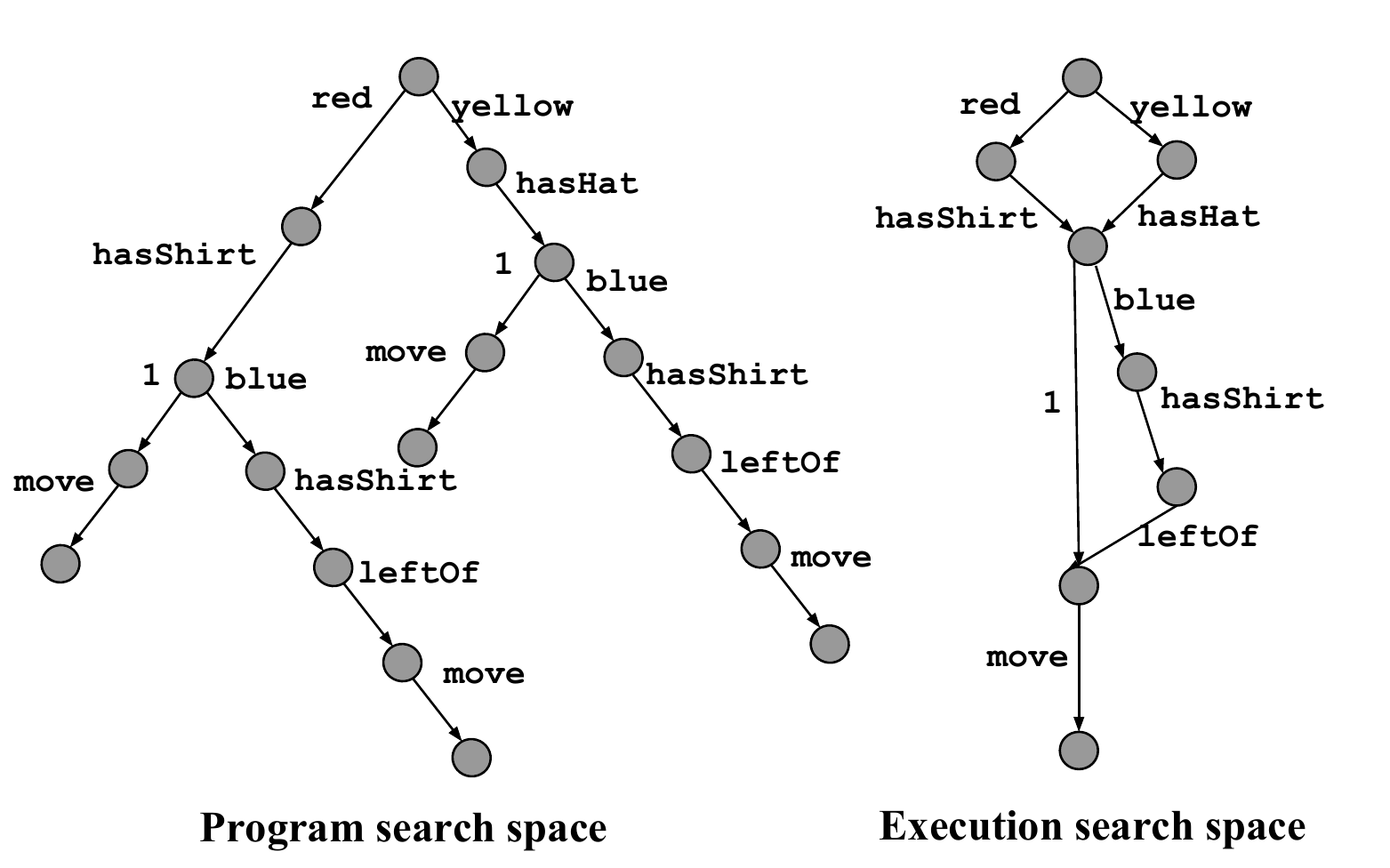}
    \caption{A set of commands represented in program space (left) and execution space (right). The commands relate to the first world in \reffig{scone_state}. Since multiple prefixes have the same execution result (e.g., \textsc{yellow hasHat} and \textsc{red hasShirt}), the execution space is smaller.}
    \label{fig:search_space}
\end{figure}

Program space can be formalized as a directed tree $T = (\sV_T, \sE_T)$, where  vertices $\sV_T$ are program prefixes, and labeled edges $\sE_T$ represent prefixes' continuations: an edge $e= (\tilde{z}, \tilde{z}')$ labeled with the token $a$, represents a continuation of the prefix $\tilde{z}$ with the token $a$, which yields the prefix $\tilde{z}'$. 
The root of the graph represents the empty sequence. Similarly, \textit{Execution space} is a directed graph $G =(\sV_G,\sE_G)$ induced from the MDP described in \refsec{mdp}. Vertices $\sV_G$ represent MDP states, which express execution results, and labeled edges $\sE_G$ represent transitions. An edge $(s_1, s_2)$ labeled by token $a$ means that $\delta(s_1, a) = s_2$.
Since multiple programs have the same execution result, execution space is a compressed representation of program space. %multiple vertices in  program space map to a single vertex in execution space. 
\reffig{search_space} shows a few commands in both program and execution space. Execution space is smaller, and so  searching in it is easier. 

Each path in execution space represents a different program prefix, and the path's final state represents its execution result. Program search can therefore be reduced to execution search: given an example $(c, \bu, y)$ and a model $\pi_{\theta}$, we can use $\pi_{\theta}$ to explore in execution space, discover \emph{correct terminal states}, i.e. states corresponding to correct full programs, and extract paths leading to those states. As the number of paths may be exponential in the size of the graph, we can use beam-search to extract the most probable correct programs (according to the model) in the discovered graph.

%We denote $G$ to be the search-graph found in the execution-search. The paths in $G$ that lead to terminal states represent the discovered programs. A program's correctness is determined by the correctness of its terminal state, i.e whether the state's world matches the target world. 

%We extract correct and incorrect programs separately. Correct programs are built with standard beam search  (guided by our model $\pi_{\theta}$) over prefixes in $G$, that their states are connected to correct terminal states. The search is therefore restricted to the space of the correct found programs. Incorrect programs are built by selecting, for each incorrect terminal state in $G$, a single path that leads to it. 

%Clearly, the advantage of this reduction is that as execution space is smaller, the search problem becomes easier. Moreover, we can score each state based on many paths that lead to that node rather than based on a single path only.%

Our approach is similar to the DPD algorithm \cite{pasupat2016inferring}, where CKY-style search is performed in denotation space, followed by search in a pruned space of programs. However, DPD was used without learning, and the search was not guided by a trained model, which is a major part of our algorithm.

\subsection{Value-based Beam Search in Execution Space}

We propose \textbf{V}alue-based \textbf{B}eam \textbf{S}earch \textbf{i}n e\textbf{X}ecution space (\ourmodel{}), a variant of beam search modified for searching in execution space, that scores search states with a value-based network. \ourmodel{} is formally defined in \refalg{vbsix}, which we will refer to throughout this section. 

Standard beam search is a breadth-first traversal of the program space tree, where a fixed number of vertices are kept in the beam at every level of the tree. The selection of vertices is done by scoring their corresponding prefixes according to $p_\theta(z_{1 \ldots t} \mid \bu, c)$. \ourmodel{} applies the same traversal in execution space (lines \ref{line:begin_for}-\ref{line:prune}). However, since each vertex in the execution space represents an execution result and not a particular prefix, we need to modify the scoring function. 

\begin{algorithm}[t]
\caption{Program Search with \ourmodel{}}
\label{alg:vbsix}
\begin{algorithmic}[1]
{\scriptsize
\Function {ProgramSearch}{$c, \bu, y, \pi_{\theta}, V_\phi$}
\State $G, \sT \leftarrow$ \Call{\ourmodel{}}{$c, \bu, y, \pi_{\theta}, V_\phi$} \label{line:call_vbsix}
\State $\sZ \leftarrow$ Find paths in $G$ that lead to states in $\sT$ with beam search\label{line:beamsearch}
\State Return $\sZ$ \label{line:return_programs}
\EndFunction
\Function {\ourmodel{}}{$c, \bu, y, \pi_{\theta}, V_\phi$}
\State $\sT \leftarrow \emptyset, P \leftarrow \{\}$ \Comment{init terminal states and DP chart}
\State $s_0 :=$ The empty program parsing state \label{line:s_zero}
\State $B_0 \leftarrow \{s_0\}, G = (\{s_0\}, \emptyset)$
\Comment{Init beam and graph}
\State $P_0[s_0] \leftarrow 1$ \Comment{The probability of $s_0$ is 1} \label{line:init_chart}
%\State $G, B, P, \sT \leftarrow \Call{Init}{}$ 
\For{$t \in [1 \ldots L]$} \label{line:begin_for}
\State $B_t \leftarrow  \emptyset$
\For{$s\in B_{t-1}, a\in \sA$}
%\For{$$}
\State $s' \leftarrow \delta(s,a)$
\State Add edge $(s, s')$ to $G$ labeled with $a$ \label{line:build_graph}
\If{$s'$ is correct terminal}
\State $\sT \leftarrow \sT \cup \{s'\}$ \label{line:build_terminal_state}
\Else
\State $P_t[s'] \leftarrow P_t[s'] + P_{t-1}[s]\cdot\pi_{\theta}(a|s)$ \label{line:update_score}
\State $B_{t} \leftarrow B_t \cup \{s'\}$ \label{line:build_beam}
\EndIf
\EndFor
\State Sort $B_t$ by \Call{AC-Scorer}{$\cdot$}
\State $B_t \leftarrow \text{Keep the top-}K \text{ states in } B_t$ \label{line:prune}
\EndFor
%\EndFor
\State Return $G, \sT$
\EndFunction
%\Function {Init}{$G, T, B, P$}
%\State $s_0 :=$ The empty program's parsing state
%\State $G \leftarrow $ Build a graph with a single vertex $s_0$  
%\State $B_0, P_0[s_0], T \leftarrow \{s_0\}, 1, \emptyset$
%\EndFunction
\Function {AC-Scorer}{$s, P_t, V_\phi, y$} \label{line:ac}
\State Return $P_t[s] + V_\phi(s, y)$ 
\EndFunction \label{line:ac_end}
}
\end{algorithmic}
\end{algorithm}

Let $s$ be a vertex discovered in iteration $t$ of the search. We propose two scores for ranking $s$. The first is \emph{the actor score}, the probability of reaching vertex $s$ after $t$ iterations\footnote{We score paths in different iterations independently to avoid bias for shorter paths. An MDP state that appears in multiple iterations will get a different score in each iteration.} according to the model $\pi_\theta$. The second and more novel score is the value-based \emph{critic score}, an estimate of the state's expected reward. The AC-Score is the sum of these two scores (lines \ref{line:ac}-\ref{line:ac_end}). 

The actor score, $p_{\theta}^t(s)$, is the sum of probabilities of all prefixes of length $t$ that reach $s$ (rather than the probability of one prefix as in beam search). \ourmodel{} approximates $p_{\theta}^t(s)$ by performing the summation only over paths that reach $s$ via states in the beam $B_{t-1}$, which can be done efficiently with a dynamic programming (DP) chart $P_t[s]$ that keeps actor score approximations in each iteration (line \ref{line:update_score}). This lower-bounds the true $p_{\theta}^t(s)$ since some prefixes of length $t$ that reach $s$ might have not been discovered. 

%\begin{equation*}\label{eq:pareto mle2}
%\begin{aligned}
%p_{\theta_t}(s) = \sum_{s'} p_{\theta_{t-1}}(s') \cdot (\sum_{\{a: \delta(s', a) = s\}}\pi_\theta(a \mid s')).
% \end{aligned}
%\end{equation*}

%\ourmodel{} approximates $p_{\theta_t}(s)$ by performing the summation over the states in the beam $B_{t-1}$ rather than all states. This can be done efficiently by maintaining a dynamic programming (DP) chart $P_t[\cdot]$ that keeps the actor scores approximations in each iteration (line \ref{line:update_score}). This is a lower bound on the true model probability for this state, since there might be prefixes of length $t$ that reach $s$ that were not discovered. 

Contrary to standard beam-search, we want to score search states also with a critic score $\E_{p_{\theta}}[R(s)]$, which is the sum of the suffix probabilities that lead from $s$ to a correct terminal state: 
$${\E}_{p_{\theta}}[R(s)] = \sum_{\tau(s)} p_\theta(\tau(s) \mid s) \cdot R(\tau(s)),$$ where $\tau(s)$ are all possible trajectories starting from $s$ and $R(\tau(s))$ is the reward observed when taking the trajectory $\tau(s)$ from $s$. Enumerating all trajectories $\tau(s)$ is intractable and so we will approximate $\E_{p_{\theta}}[R(s)]$ with a trained value network $V_\phi(s, y)$, parameterized by $\phi$. 
Importantly, because we are searching at training time, we can condition $V_\phi$ on both the current state $s$ and target denotation $y$. At test time we will use $\pi_\theta$ only, which does not need access to $y$. 

Since the value function and DP chart are used for efficient ranking, the asymptotic run-time complexity of \ourmodel{} is the same as standard beam search ($\sO(K \cdot |\sA| \cdot L$)). The beam search in Line~\ref{line:beamsearch}, where we extract programs from the constructed execution space graph, can be done with a small beam size, since it operates over a small space of correct programs. Thus, its contribution to the algorithm complexity is negligible. 

\begin{figure}
    \centering
    \includegraphics[width=6.5cm, height=4.5cm]{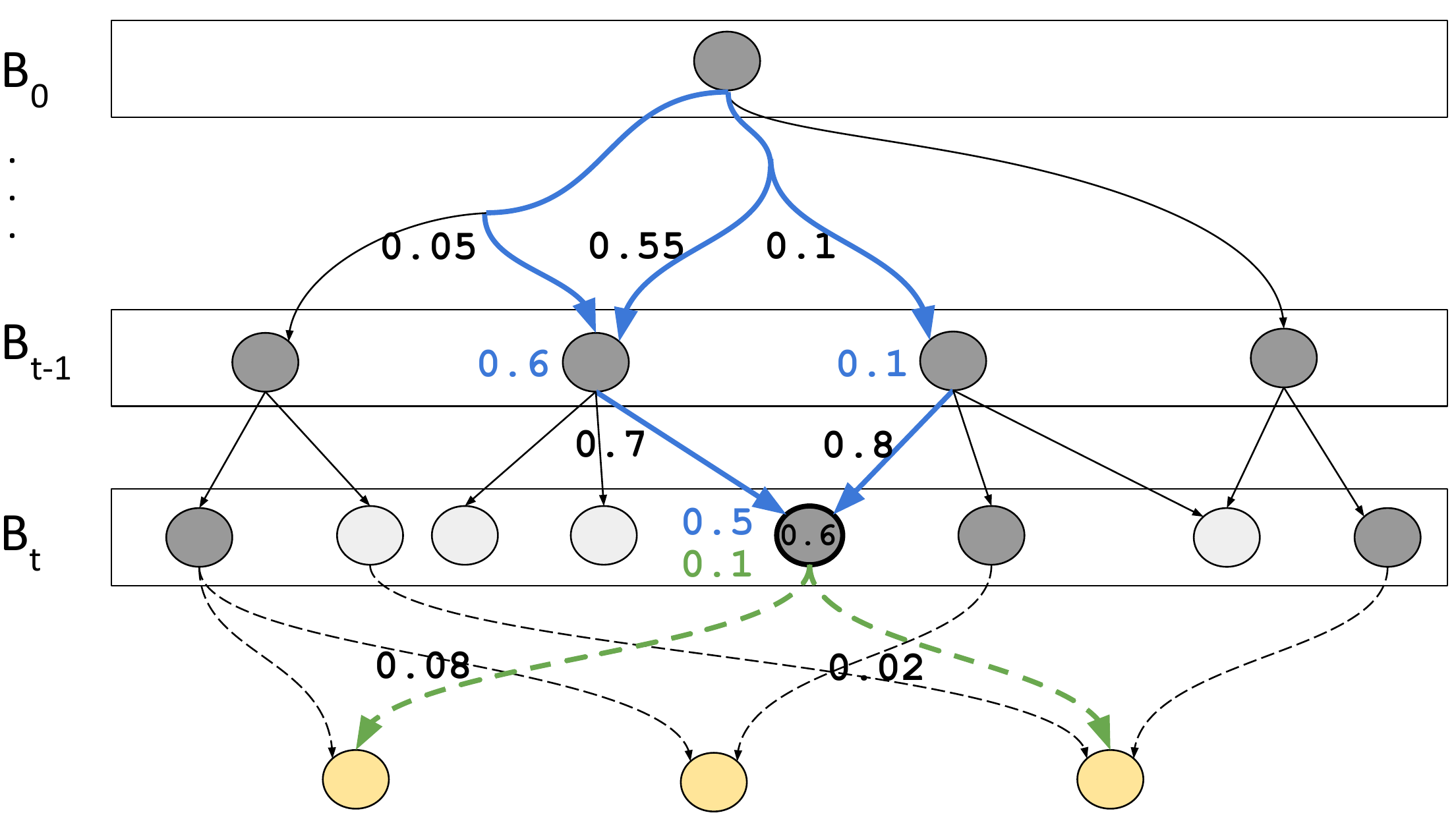}
    \caption{An illustration of scoring and pruning states in step $t$ of \ourmodel{} (see text for details). Discovered edges are in full edges, and undiscovered edges are dashed, correct terminal states are in yellow, states that are kept in the beam are in dark grey, actor scores are in blue, and critic scores in green.}
    \label{fig:exbs}
\end{figure}

\reffig{exbs} visualizes scoring and pruning with the actor-critic score in iteration $t$. Vertices in $B_t$ are discovered by expanding vertices in $B_{t-1}$, and each vertex is ranked by the AC-scorer. The highlighted vertex score is $0.6$, a sum of the actor score ($0.5$) and the critic score ($0.1$). The actor score is the sum of its prefixes ($(0.05 + 0.55) \cdot 0.7 + 0.01 \cdot 0.08 = 0.5$) and the critic score is a value network estimate for the sum of probabilities of outgoing trajectories reaching correct terminal states ($0.02 + 0.08 = 0.1$). Only the top-$K$ states are kept in the beam ($K=4$ in the figure).

\ourmodel{} leverages execution space in a number of ways. First, since each vertex in execution space compactly represents multiple prefixes, a beam in \ourmodel{} effectively holds more prefixes than in standard beam search. Second, running beam-search over a graph rather than a tree is less greedy, as the same vertex can surface back even if it fell out of the beam. 

The value-based approach has several advantages as well: First, evaluating the probability of outgoing trajectories provides look-ahead that is missing from standard beam search. Second (and most importantly), $V_\phi$ is conditioned on $y$, which $\pi_\theta$ doesn't have access to, which allows finding correct programs with low model probability, that $\pi_\theta$ can learn from.
This resembles the actor-critic model proposed by \cite{bahdanau2017actor}, but they assumed a fully-supervised setup where the output sequence is observed.
We note that our two contributions are orthogonal: the critic score can be used in program space, and \ourmodel{} can use the actor score only.

%% file: 05_training.tex
\section{Training}\label{sec:training}

We train the model $\pi_{\theta}$ and value network $V_{\phi}$ jointly (\refalg{ac-train}). $\pi_{\theta}$ is trained using MML over discovered correct programs (Line~\ref{line:return_programs}, Algorithm~\ref{alg:vbsix}). The value network is trained as follows: Given a training example $(c, \bu, y)$, we generate a set of correct programs $\sZ_\text{pos}$ with \ourmodel. The value network needs negative examples, and so for every incorrect terminal state $z_{\text{neg}}$ found during search with \ourmodel{} we create a single program leading to $z_{\text{neg}}$. We then construct 
a set of training examples $\sD_v$, where each example $((s, y), l)$ labels states encountered while generating programs $z \in \sZ$ with the probability mass of correct programs suffixes that extend it, i.e., 
$l = \sum_{z_t} p_{\theta}(z_{t\ldots |z|})$, where $z_t$ ranges over all $z \in \sZ$ and $t \in [1 \ldots |z|]$
%$l = \sum_{\{z \in \sZ, t \in [1 \ldots |z|] : \denp{z_{1\ldots t}}{c} = s, R(z) = 1\}} p_{\theta}(z_{t\ldots |z|})$.
Finally, we train $V_{\phi}$ to minimize the log-loss objective:
{
\medmuskip=-1mu
\thinmuskip=-1mu
\thickmuskip=-1mu
$\sum_{((s, y), l) \in \sD_v} l\cdot\log{V_\phi(s, y)} + (1-l)(\log{(1 - V_\phi(s, y))})$
}.

Similar to actor score estimation, labeling examples for $V_\phi$ is affected by beam-search errors: the labels lower bound the true expected reward. However, since search is guided by the model, those programs are likely to have low probability.
Moreover, estimates from $V_\phi$ are based on multiple examples, compared to probabilities in the DP chart, and are more robust to search errors.
%DP chart for one particular example. This makes the value network more robust to beam search errors.

\begin{algorithm}[t]
\caption{Actor-Critic Training}
\label{alg:ac-train}
\begin{algorithmic}[1]
{\scriptsize
\Procedure {Train}{\null}
\State Initialize $\theta$ and $\phi$ randomly 
\While{$\pi_{\theta} $ not converged}
\State $(x:=(c, \bu), y) \leftarrow $ select random example
\State $\sZ_\text{pos} \leftarrow \Call{ProgramSearch}{c, \bu, y, \pi_{\theta}, V_{\phi}}$ \label{line:prog_search}
\State $\sZ_{\text{neg}} \leftarrow$ programs leading to incorrect terminal states
\State $\sD_v \leftarrow \Call{BuildValueExamples}{(\sZ_\text{pos} \cup \sZ_\text{neg}), c, y}$ \label{line:build_value_examples}
%\State $V_\phi \leftarrow \Call{Fit}{V_\theta, E_V}$%
\State Update $\phi$ using $\sD_v$, update $\theta$ using $(x, \sZ_\text{pos}, y)$
%\State Update $\theta$ with MML 
%\State $\pi_{\theta} \leftarrow \Call{Up-Weight}{\pi_\theta, Z^+}$%
\EndWhile 
\EndProcedure
\Function {BuildValueExamples}{$\sZ, c, \bu, y$}
\For{$z \in \sZ$}
\For{$t \in [1 \ldots |z|]$}
\State $s \leftarrow \denp{z_{1\ldots t}}{c}$
\State $L[s] \leftarrow L[s] + p_{\theta}(z_{t\ldots |z|} \mid c, \bu)\cdot R(z)$
\EndFor
\EndFor
\State $\sD_v \leftarrow \{((s, y), L[s])\}_{s \in L}$
\State Return $\sD_v$
\EndFunction
}
\end{algorithmic}
\end{algorithm}

\paragraph{Neural network architecture: } We adapt the  model proposed by \newcite{guu2017bridging} for SCONE. The model receives the current utterance $u^i$ and program stack $\psi$, and returns a distribution over the next token. Our value network receives the same input, but also the next utterance $u^{i+1}$, the world state $w^i$ and target world state $y$, and outputs a scalar. Appendix \ref{sec:nn} provides a full description.

%% file: 06_experiments.tex
\section{Experiments}
\begin{table*}[t]
\begin{center}
{\scriptsize
\hfill{}
\begin{tabular}{l|l|l|l|l|l|l|l|l}
& & \multicolumn{2}{|c|}{\textbf{\scenedomain{}}} &
\multicolumn{2}{|c|}{\textbf{\alchemydomain{}}} &
\multicolumn{2}{|c}{\textbf{\tangramdomain{}}} \\
 & \textbf{Train beam size}& \textbf{3 utt} & \textbf{5 utt} & \textbf{3 utt} & \textbf{5 utt} & \textbf{3 utt} & \textbf{5 utt} \\
\hline
%\reinforce{} & 0$\pm(0.0)$ & 0$\pm(0.0)$ & 0$\pm(0.0)$ & 0$\pm(0.0)$ & 0$\pm(0.0)$ & 0$\pm(0.0)$  \\
\mml{}  & 32 & 8.4$\pm(2.0)$ & 7.2$\pm(1.3)$ & 41.9$\pm(22.8)$ & 33.2$\pm(20.1)$ & 32.5$\pm(20.7)$ & 16.8$\pm(14.1)$ \\
& 64 & 15.4$\pm(12.6)$ & 12.3$\pm(9.6)$ & 44.6$\pm(23.7)$ & 36.3$\pm(20.6)$ & 45.6$\pm(18.0)$ & 25.8$\pm(12.6)$ \\ \cline{1-8}
\expertmml{} & 32 & 1.8$\pm(1.5)$ &  1.6$\pm(1.2)$ & 29.4$\pm(22.7)$ & 23.1$\pm(18.8)$ & 2.4$\pm(0.8)$ & 1.2$\pm(0.5)$ \\  
\ourmodel{}  & 32 & \textbf{34.2$\pm(27.5)$} & \textbf{28.2$\pm(20.7)$} & \textbf{74.5}$\pm(1.1)$ & \textbf{64.8}$\pm(1.5)$ &  \textbf{65.0}$\pm(0.8)$ & \textbf{43.0$\pm(1.3)$} \\ 
\end{tabular}}
\hfill{}
\caption{ 
Test accuracy and standard deviation of \ourmodel{} compared to \mml{} baselines (top) and our training methods (bottom). We evaluate the same model over the first 3 and 5 utterances in each domain.
}
\label{baselines}
\end{center}
\end{table*}

\begin{table*}[t]
\begin{center}
{\scriptsize
\hfill{}
\begin{tabular}{l|l|l|l|l|l|l|l|l}
&  & \multicolumn{2}{|c|}{\textbf{\scenedomain}} &
\multicolumn{2}{|c|}{\textbf{\alchemydomain{}}} &
\multicolumn{2}{|c}{\textbf{\tangramdomain{}}} \\
 \textbf{Search space} & \textbf{Value} &\textbf{3 utt} & \textbf{5 utt} & \textbf{3 utt} & \textbf{5 utt} & \textbf{3 utt} & \textbf{5 utt} \\
\hline
Program & No & 5.5$\pm(0.5)$  & 3.8$\pm(0.6)$ & 36.4$\pm(26.5)$ & 25.4$\pm(23.0)$  & 34.4$\pm(18.3)$  & 15.6$\pm(12.8)$ \\
Execution & No & 7.4$\pm(10.4)$ & 4.0$\pm(5.8)$ & 41.3$\pm(28.6)$ & 28.2$\pm(23.27)$ & 33.5$\pm(15.5)$ & 12.7$\pm(10.4)$ \\
Program & Yes &  7.6$\pm(8.3)$ & 3.4$\pm(2.9)$ & 78.5$\pm(1.0)$ & 72.8$\pm(1.3)$ &   66.8$\pm(1.5)$ & 42.8$\pm(1.9)$  \\
Execution & Yes & \textbf{31.0}$\pm(24.7)$ & \textbf{22.6}$\pm(19.6)$ & \textbf{81.9}$\pm(1.3)$ & \textbf{75.2}$\pm(2.9)$ & \textbf{68.6}$\pm(2.0)$ & \textbf{44.2}$\pm(2.1)$ \\
\end{tabular}}
\hfill{}
\caption{ 
Validation accuracy when ablating the different components of \ourmodel. The first line presents \mml{}, the last line is \ourmodel{}, and the intermediate lines examine execution space and value-based networks separately.
}
\label{tab:ablations}
\end{center}
\end{table*}

\subsection{Experimental setup} 
We evaluated our method on the three domains of SCONE with the standard accuracy metric, i.e., the proportion of test examples where the predicted program has the correct denotation. We trained with \ourmodel{}, and used standard beam search ($K=32$) at test time for programs' generation. Each test example contains 5 utterances, and similar to prior work we reported the model accuracy on all 5 utterances as well as the first 3 utterances. We ran each experiment 6 times with different random seeds and reported the average accuracy and standard deviation.

In contrast to prior work on SCONE \cite{long2016projections,guu2017bridging,Suhr2018Situated}, where models were trained on all sequences of 1 or 2 utterances, and thus were exposed during training to all gold intermediate states, we trained from longer sequences keeping intermediate states latent.
This leads to a harder search problem that was not addressed previously, but makes our results incomparable to previous results \footnote{For completeness, we show the performance on these datasets from prior work in Appendix \ref{sec:prior}.}.
%Unlike previous works in SCONE, where the model was trained from  instruction sequences of 1 and 2 utterances  \cite{long2016projections,guu2017bridging,Suhr2018Situated}, we trained the model from longer instruction sequences. 
In \scenedomain{} and \tangramdomain{}, we used 
the first 4 and 5 utterances as examples. In \alchemydomain{}, we used the first utterance and 5 utterances.

%sequences of 4 and 5 utterances, and in the Alchemy domain we used sequences of 1 and 5 utterances. We note that the shorter examples are the first instructions of the full-5 utterances examples. 
%This experimental setting raises a challenging search problem in training that previous works have not dealt with. 

\paragraph{Training details}
To warm-start the value network, we trained it for a few thousand steps, and only then start re-ranking with its predictions. Moreover, we gain efficiency by first returning $K_0$(=128) states with the actor score, and then re-ranking with the actor-critic score, returning $K$(=32) states. Last, we use the value network only in the last two utterances of every example since we found it has less effect in earlier utterances where future uncertainty is large. 
We used the Adam optimizer \cite{kingma2014adam} and fixed GloVe embeddings \cite{pennington2014glove} for utterance words.
%Fixed GloVe vectors \cite{pennington2014glove} are used to embed utterance words, and all other parameters are initialized randomly. The Adam optimizer \cite{kingma2014adam} was used to train the models, with  learning rate of 0.001 and a mini-batch size of 8 examples.

%Number of implementation choices were made regaring the value network: (a) The value network is trained only from states that relate to the last two utterances of the examples, and it is only use to rank such states. (b) The value network is used for ranking only after a fixed number training steps (c) Ranking is done by sorting all the possible continuations according to their actor-score, and then re-ranking only the top states with the actor-critic scorer. This allows us to control the value network's input batch size.            
\begin{figure*}
    \centering
    \includegraphics[width=16cm, height=4cm]{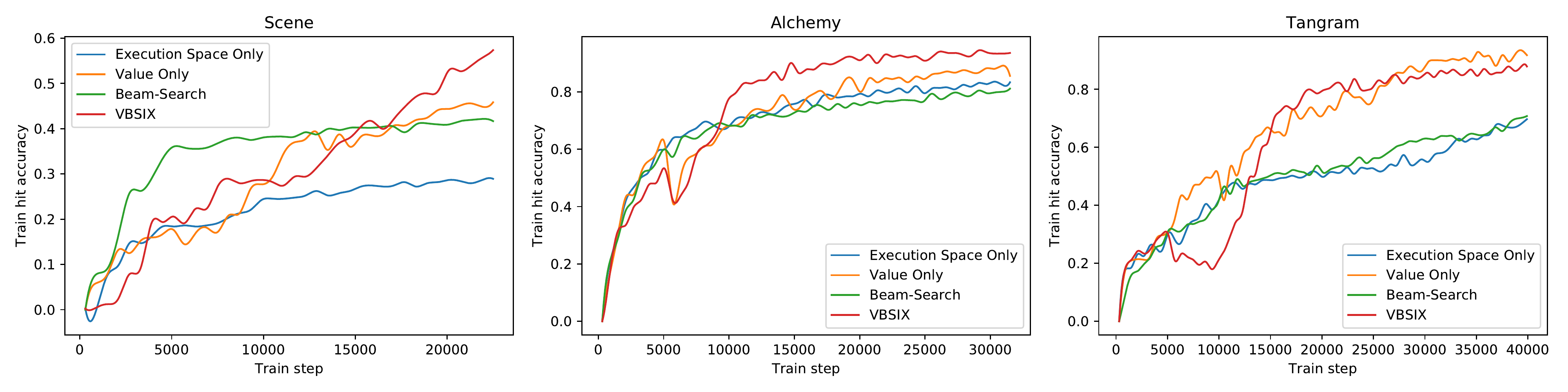}
    \caption{
    Training hit accuracy on examples with 5 utterances, comparing \ourmodel{} to baselines with ablated components. The results are averaged over 6 runs with different random seeds.}
    \label{fig:hit_train}
\end{figure*}

\begin{figure*}
    \centering
    \includegraphics[width=16cm, height=4cm]{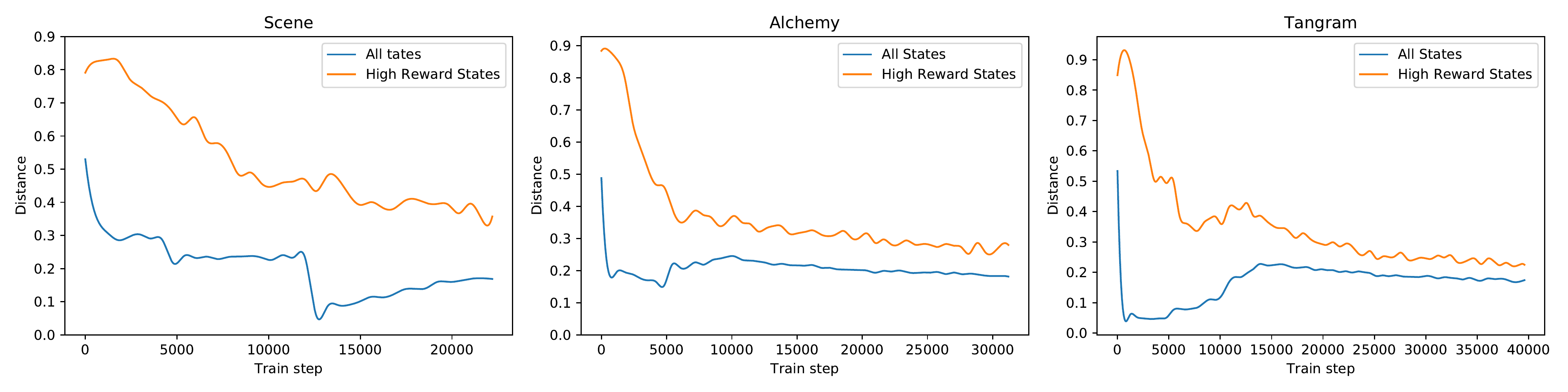}
    \caption{
    The difference between the prediction of the  value network and the expected reward (estimated from the discovered paths) during training. We report the average distance for all of the states (blue) and for the high reward states only ($>0.7$, orange). The results are averaged over 6 runs with different random seeds.}
    \label{fig:value_train}
\end{figure*}

\paragraph{Baselines} 
We evaluated the following training methods (Hyper-parameters are in appendix~\ref{sec:hyper}): 
\begin{enumerate}[wide,labelindent=0pt,topsep=0pt,noitemsep]
\item \mml{}: Our main baseline, where search is done with beam search and training with MML. We used randomized beam-search,
which adds $\epsilon$-greedy exploration to beam search, 
which was proposed by \newcite{guu2017bridging} and performed better \footnote{We did not include meritocratic updates \cite{guu2017bridging}, since it performed worse in initial experiments.}.
%\footnote{We did not include meritocratic updates \cite{guu2017bridging}, since it performed worse in initial experiments.}
%\item \reinforce{} \cite{williams1992simple, sutton1999policy}: We followed the implementation of \newcite{guu2017bridging}, which includes variance reduction with a constant baseline and  $\epsilon$-greedy exploration.
\item \expertmml{}: An alternative way of using the target denotation $y$ at training time, based on imitation learning \cite{daume09searn,ross2011reduction,berant2015agenda}, is to train an \emph{expert policy} $\pi_\theta^{\text{expert}}$, which receives $y$ as input in addition to the parsing state, and trains with the MML objective. Then, our policy $\pi_\theta$ is trained using programs found by $\pi_\theta^{\text{expert}}$. The intuition is that the expert can use $y$ to find good programs that the policy $\pi_\theta$ can train from.
\item \ourmodel{}: Our proposed training algorithm. %that ranks states with an actor-critic score and searches in execution space.
\end{enumerate}
We also evaluated \reinforce{}, where Monte-Carlo sampling is used as search strategy \cite{williams1992simple, sutton1999policy}. We followed the implementation of \newcite{guu2017bridging}, who performed variance reduction with a constant baseline and added $\epsilon$-greedy exploration. We found that \reinforce{} fails to discover any correct programs to bootstrap from, and thus its performance was very low.

\subsection{Results}

Table \ref{baselines} reports test accuracy of \ourmodel{} compared to the baselines. First, \ourmodel{} outperforms all baselines in all cases. \mml{} is the strongest baseline, but even with an increased beam ($K=64$), \ourmodel{} ($K=32$) surpasses it by a large margin (more than 20 points on average). On top of the improvement in accuracy, in \alchemydomain{} and \tangramdomain{} the standard deviation of \ourmodel{} is lower than the other baselines across the 6 random seeds, showing the robustness of our model. %One exception is the \scenedomain{} domain, where the language is more complex and absolute accuracies are lower. The performance of other baselines is quite low and consequently also the standard deviation.

\expertmml{} performs worse than \mml{} in all cases. We hypothesize that using the denotation $y$ as input to the expert policy $\pi_\theta^\text{expert}$ results in many spurious programs, i.e., they are unrelated to the utterance meaning. This is since the expert can learn to perform actions that take it to the target world state while ignoring the utterances completely. Such programs will lead to bad generalization of $\pi_\theta$. Using a critic at training time eliminates this problem, since its scores depend on $\pi_\theta$.

%across the board. In particular, on \alchemydomain{} \tangramdomain{} \ourmodel{} achieves significant improvement over standard-beam search. In addition, its standard deviation is much lower, which reflect the algorithm robustness against the effects of randomness. However, in the \scenedomain{} domain the \ourmodels accuracy improvement is lower. This is due to the fact that \scenedomain{} has the most complex language, and therefore poses the hardest search problem. 

%Expert-MML achieved lower accuracy results on all domains compared to the standard MML and \ourmodel, with the exception of \tangramdomain{}'s 5-utterances examples test. The main reason for Expert-MML poor performance is the effect of spurious programs - programs that produce the correct target world but do not reflect the semantics of the instructions. The expert search policy can learn how to generate such spurious programs given the target denotation. The model that only has access to the utterances, which is used at test time, can't generalize from such examples. In \tangramdomain{}, the fact that the accuracy score of the 5-utterances examples is much higher than the accuracy score of the first 3 utterances examples, implies that many of them are spurious. The model managed to learn a strategy to generate spurious programs in test-time. We note that this issue doesn't accrue in \ourmodel, since the value network estimates the expected reward, a metric that relies on the model's distribution, which is a function of the utterances.  

\paragraph{Ablations} 
We performed ablations to examine the benefit of our two technical contributions (a) execution space (b) value-based search. \reftab{ablations} presents accuracy on the validation set when each component is used separately, when both of them are used (\ourmodel), and when none are used (beam-search). We find that both contributions are important for performance, as the full system achieves the highest accuracy in all domains. In \scenedomain{}, each component has only a slight advantage over beam-search, and therefore both are required to achieve significant improvement. However, in \alchemydomain{} and \tangramdomain{} most of the gain is due to the value network.  

We also directly measured the \emph{hit accuracy} at training time, i.e., the proportion of training examples where the beam produced by the search algorithm contains a program with the correct denotation.
This measures the effectiveness of search at training time. In \reffig{hit_train}, we show train hit accuracy in each training step, averaged across 6 random seeds. The graphs illustrate the performance of each search algorithm in every domain throughout training.
We observe that validation accuracy results are correlated with the improvement in hit accuracy, showing that better search leads to better overall performance.

\subsection{Analysis}

\paragraph{Execution Space} 
We empirically measured two quantities that we expect should reflect the advantage of execution-space search. First, 
we measured the number of programs stored in the execution space graph compared to beam search, which holds $K$ programs.
%to evaluate the amount of comp program-space compression, we measured the effective number of paths a beam holds on average, %i.e the number of discovered paths that lead to the beam's states. 
Second, we counted the average number of states that are connected to correct terminal states in the discovered graph, but fell out of the beam during  search. The property reflects the gain from running search over a graph structure, where the same vertex can resurface. We preformed the analysis on \ourmodel{} over 5-utterance training examples in all 3 domains. The following table summarizes the results:
\begin{table}[h]
\begin{center}
{\scriptsize
\hfill{}
\begin{tabular}{l|l|l|l}
 \textbf{Property} & \textbf{\scenedomain} & \textbf{\alchemydomain} & \textbf{\tangramdomain}  \\
\hline
Paths in beam & 143903 & 5892 & 678 \\
Correct discarded states & 18.5 & 11.2 & 3.8 \\
\end{tabular}}
\hfill{}
\label{tab:graph}
\end{center}
\end{table}

We found the measured properties and the contribution of execution space in each domain are correlated, as seen in the ablations. Differences between domains are due to the different complexities of their formal languages. As the formal language becomes more expressive, the execution space is more compressed as each state can be reached in more ways. In particular, the formal language in \scenedomain{} contains more functions compared to the other domains, and so it benefits the most from execution-space search.

\paragraph{Value Network} We analyzed the accuracy of the value network at training time by measuring, for each state, the difference between its expected reward (estimated from the discovered paths) and the value network prediction. \reffig{value_train} shows the average difference in each training step for all encountered states (in blue), and for high reward states only (states with expected reward larger than $0.7$, in orange). Those metrics are averaged across 6 runs with different random seeds. 

The accuracy of the value network improves during training, except when the policy changes substantially, in which case the value network needs to re-evaluate the policy. When the value network converges, the difference between its predictions and the expected reward is $0.15-0.2$ on average. However, for high reward states the difference is higher ($\sim 0.3$). This indicates that the value network has a bias toward lower values, which is expected as most states lead to low rewards. Since \ourmodels uses the value network as a beam-search ranker, the value network doesn't need to be exact as long as it assigns higher values to states with higher expected rewards. Further analysis of the value network is also provided in appendix~\ref{sec:vna}. 

%\paragraph{Value Network} We analyzed the ability of the value network to predict expected reward. We collected a random set of 120 value-network examples in \scenedomain{}. We found that the features that effect the value network predictions the most are: (a) the number of differences between the current world and the target world, which implies whether the correct terminal state is reachable, and (b) the number of people in the state's world, which relates to the example's complexity. We give more details on this analysis in appendix~\ref{sec:vna}. %

%% file: 07_related.tex
\section{Related Work}
Training from denotations has been extensively investigated \cite{kwiatkowski2013scaling,pasupat2015compositional,bisk2016natural}, with a recent emphasis on neural models \cite{neelakantan2016neural,krishnamurthy2017neural}. Improving beam search has been investigated by proposing specialized objectives \cite{wiseman2016beam}, stopping criteria \cite{yang2018breaking}, and using continuous relaxations \cite{goyal2018continuous}.

\newcite{bahdanau2017actor} and \newcite{Suhr2018Situated} proposed ways to evaluate intermediate predictions from a sparse reward signal. \newcite{bahdanau2017actor} used a critic network to estimate expected BLEU in translation, while \newcite{Suhr2018Situated} used edit-distance between the current world and the goal for SCONE. But, in those works stronger supervision was assumed: \newcite{bahdanau2017actor} utilized the gold sequences, and \newcite{Suhr2018Situated} used intermediate worlds states. Moreover, intermediate evaluations were used to compute gradient updates, rather than for guiding search.

Guiding search with both policy and value networks was done in Monte-Carlo Tree Search (MCTS) for tasks with a sparse reward \cite{silver2017mastering, anthony2017thinking, shen2018reinforcewalk}. In MCTS, value network evaluations are refined with backup updates to improve policy scores. In this work, we gain this advantage by using the target denotation. The use of an actor and a critic is also reminiscent of $A^*$ where states are scored by past cost and an admissible heuristic for future cost \cite{klein2003fast, pauls2009kbest, lee2016global}. In semantic parsing, \newcite{misra2018policy} recently proposed a critic distribution to improve the policy. However, their proposed critic is based on prior domain knowledge, while in our work the critic is a learned parameterized function. 

%Using the target output to train a critic at training time has been suggested by \newcite{bahdanau2017actor}, but assuming a  supervised setup where the full output sequence is observed rather than denotation. \newcite{Suhr2018Situated} used the target world state of SCONE to manually-define a reward function for a reinforcement learning algorithm, but did not learn a model that considers the denotation and \jb{say how we are different}.

%Understanding instructions has been studied with the SAIL corpus \cite{macmahon2006walk,chen11navigate,andreas2015alignment,mei2016listen} and recently with other environments, focusing on single sentence instructions \cite{janner2018representation,misra2018mapping,anderson2018vision,tan2018source}.

%Tackling the limitations of beam search has been investigated recently by proposing objectives suitable for beam search \cite{wiseman2016beam}, by using continuous relaxation that afford differentiability \cite{goyal2018continuous}, and by developing specialized stopping criteria \cite{yang2018breaking}.

%beam search stuff

%% file: 08_conclusions.tex
\section{Conclusions}
In this work, we propose a new training algorithm for mapping instructions to programs given denotation supervision only. Our algorithm exploits the denotation at training time to train a critic network used to rank search states on the beam, and performs search in a compact execution space rather than in the space of programs. We evaluated on three different domains from SCONE, and found that it dramatically improves performance compared to strong baselines across all domains.

\ourmodel{} is applicable to any task that supports graph-search exploration. Specifically, for tasks that can be formulated as an MDP with a deterministic transition function, which allow efficient execution of multiple partial trajectories. Those tasks include a wide range of instruction mapping  \cite{branavan09reinforcement, vogel10navigate, anderson2018vision} and semantic parsing tasks \cite{dahl1994expanding, iyyer2017search, yu2018spider}. Therefore, evaluating \ourmodel{} on other domains is a natural next step for our research.

\section*{Acknowledgments}
We thank the anonymous reviewers for their constructive
feedback. This work was completed in fulfillment for the M.Sc degree of the first author. This research was partially supported by The Israel Science Foundation grant 942/16, the Blavatnik Computer Science Research Fund, and The Yandex Initiative for Machine Learning.

%% file: 09_nn.tex
\clearpage
\section{Neural Network Architecture}\label{sec:nn}

We adopt the model $\pi_\theta(\cdot)$ proposed by \newcite{guu2017bridging}. The model receives the current utterance $u^i$ and the program stack $\psi$. A bidirectional LSTM \cite{hochreiter1997lstm} is used to embed $u_i$, while $\psi$ is embedded by concatenating the embedding of stack elements. The embedded input is then fed to a feed-forward network with attention over the LSTM hidden states, followed by a softmax layer that predicts a program token. 
Our value network $V_\phi(\cdot)$ shares the input layer of $\pi_\theta(\cdot)$. In addition, it receives the next utterance $u^{i+1}$, the current world state $w^i$ and the target world state $w^M$. The utterance $u^{i+1}$ is embedded with an additional BiLSTM, and world states are embedded by concatenating embeddings of SCONE elements. The inputs are concatenated and fed to a feed-forward network, followed by a sigmoid layer that outputs a scalar.        
\begin{figure}
    \centering
    \includegraphics[width=7.7cm, height=6.2cm]{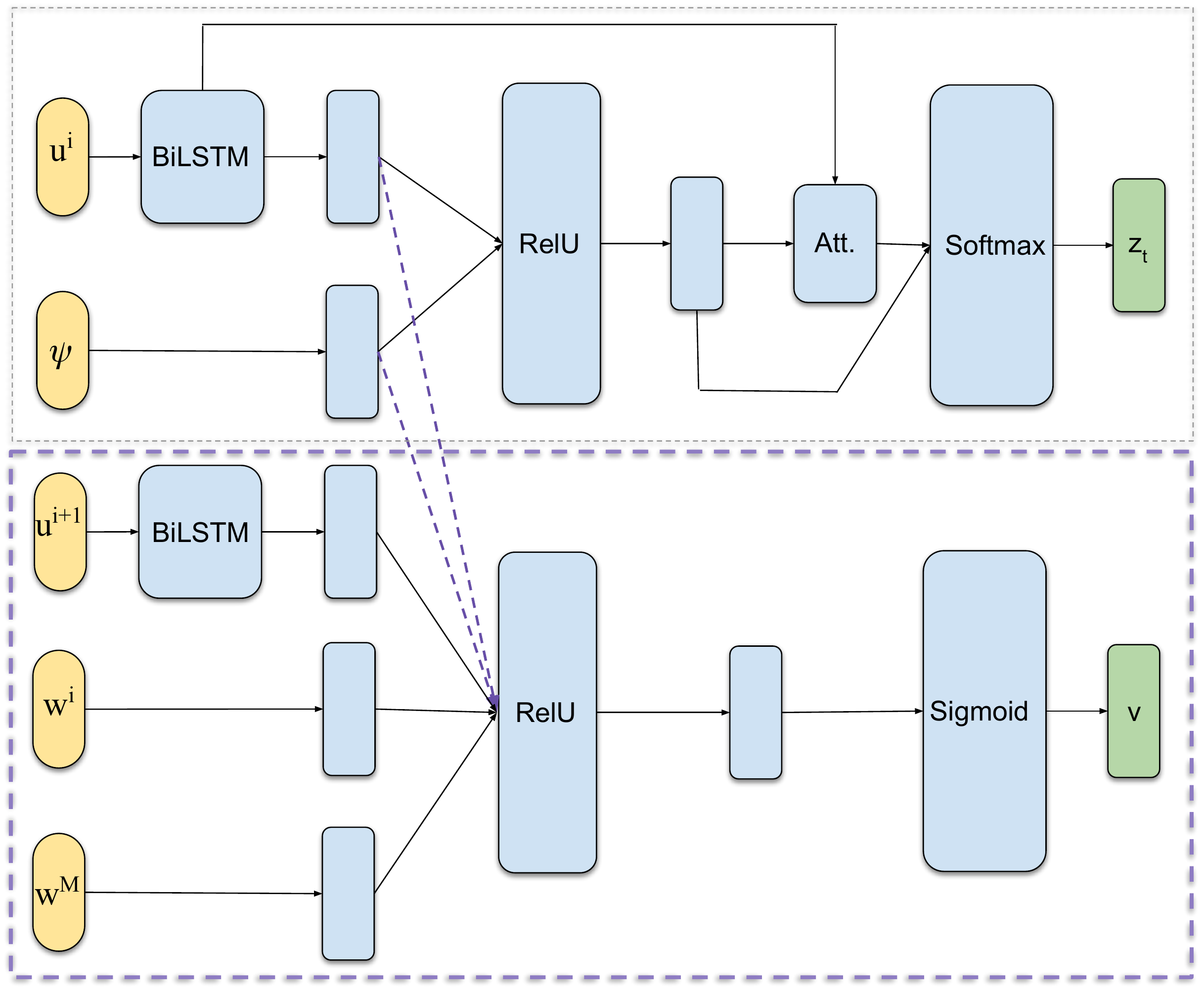}
    \caption{The model proposed by \newcite{guu2017bridging} (top), and our value network (bottom).}
    \label{fig:network}
\end{figure}

%% file: 10_hyper.tex
\section{Hyper-parameters}\label{sec:hyper}
\reftab{hyper} contains the hyper-parameter setting for each experiment.  Hyper-parameters of \reinforce{} and \mml{} were taken from \newcite{guu2017bridging}. In all experiments learning rate was 0.001 and mini-batch size was 8. We explicitly define the following hyper-parameters which are not self-explanatory:
\begin{enumerate}[wide,labelindent=0pt,topsep=0pt,noitemsep]
\item \emph{Training steps}: The number of training steps taken. 
\item \emph{Sample size}: Number of samples drawn from $p_\theta$ in \reinforce
\item \emph{Baseline}: A constant subtracted from the reward for variance reduction.
\item \emph{Execution beam size}: $K$ in Algorithm~\ref{alg:vbsix}.
\item \emph{Program bea size}: Size of beam in line~\ref{line:beamsearch} of Algorithm~\ref{alg:vbsix}.
\item \emph{Value ranking start step}: Step when we start ranking states using the critic score.
\item \emph{Value re-rank size}: Size of beam $K_0$ returned by the actor score before re-ranking with the actor-critic score.
\end{enumerate}

%% file: 11_prev.tex
\section{Prior Work}\label{sec:prior}
In prior work on SCONE, models were trained on sequences of 1 or 2 utterances, and thus were exposed during training to all gold intermediate states \cite{long2016projections,guu2017bridging,Suhr2018Situated}. \newcite{fried2018unified} assumed access to the full annotated logical form. In contrast, we trained from longer sequences, keeping the logical form and intermediate states latent. 
We report the test accuracy as reported by prior work and in this paper, noting that our results are an average of 6 runs, while prior work reports the median of 5 runs.

Naturally, our results are lower compared to prior work that uses much stronger supervision. This is because our setup poses a hard search problem at training time, and also requires overcoming \emph{spuriousness} -- the fact that even incorrect programs sometimes lead to high reward.

%In several tasks, training a policy with a strong supervision is comparable to on-policy reinforcement learning algorithms with a weaker supervision, as training with strong supervision is easier, while on the other hand the trained policy might not generalize as well, due to mismatch between the states' distribution at test and train time. 

%However, in semantic parsing tasks (such as SCONE), strong supervision also eliminates the spuriousness problem - learning from programs that are consistent with the supervision, but don't express the semantics of the natural language input. Therefore, strong supervision usually leads to better generalization in such tasks. The setting in our work is more challenging compare to prior works in SCONE, as it both poses a hard search problem in train time, and adds more noise to the reward signal.  

\begin{table}[h]
\begin{center}
{\scriptsize
\hfill{}
\begin{tabular}{l|l|l|l|l|l|l|l}
&  \multicolumn{2}{|c|}{\textbf{\scenedomain}} &
\multicolumn{2}{|c|}{\textbf{\alchemydomain{}}} &
\multicolumn{2}{|c}{\textbf{\tangramdomain{}}} \\
  &\textbf{3 utt} & \textbf{5 utt} & \textbf{3 utt} & \textbf{5 utt} & \textbf{3 utt} & \textbf{5 utt} \\
\hline
\newcite{long2016projections} & 23.2 & 14.7 & 56.8 & 52.3 & 64.9  & 27.6 \\
\newcite{guu2017bridging} & 64.8 & 46.2 & 66.9 & 52.9 & 65.8 & 37.1 \\
\newcite{Suhr2018Situated} & 73.9 & 62.0 & 74.2 & 62.7 & 80.8 & 62.4 \\ \cline{1-7}
\newcite{fried2018unified} & -- & 72.7 & -- & 72.0 & -- & 69.6 \\
\cline{1-7}
\ourmodel & 34.2 & 28.2 & 74.5 & 64.8 &  65.0 & 43.0 \\
\end{tabular}}
\hfill{}
\caption{ 
Test accuracy comparison to prior work. 
}
\label{tab:prior}
\end{center}
\end{table}

\begin{table*}[t]
\begin{center}
{\scriptsize
\hfill{}
\begin{tabular}{l|l|l|l}
 \textbf{System} & \textbf{\scenedomain{}}& \textbf{\alchemydomain{}} & \textbf{\tangramdomain{}} \\
\hline
\text{REINFORCE} & $\text{Training steps}=22.5k$ & $\text{Training steps}=31.5k$ & $\text{Training steps}=40k$ \\ 
& $\text{Sample size}=32$ & $\text{Sample size}=32$ & $\text{Sample size}=32$ \\ 
 & $\epsilon=0.2$ & $\epsilon=0.2$ & $\epsilon=0.2$ \\
 & $\text{Baseline}=10^{-5}$ & $\text{Baseline}=10^{-2}$ & $\text{Baseline}=10^{-3}$ \\ 
 \cline{1-4}
\text{MML} & $\text{Training steps}=22.5k$ & $\text{Training steps}=31.5k$ & $\text{Training steps}=40k$ \\
& $\text{Beam size}=32$ & $\text{Beam size}=32$ & $\text{Beam size}=32$ \\ 
 & $\epsilon=0.15$ & $\epsilon=0.15$ & $\epsilon=0.15$ \\
 \cline{1-4} 
\text{VBSiX}  & $\text{Training steps}=22.5k$ & $\text{Training steps}=31.5k$ & $\text{Training steps}=40k$ \\
& $\text{Execution beam size}=32$ & $\text{Execution beam size}=32$ & $\text{Execution beam size}=32$ \\ 
 & $\text{Program beam size}=8$ & $\text{Program beam}=8$ & $\text{Program beam size}=8$ \\
 & $\epsilon=0.15$ & $\epsilon=0.15$ & $\epsilon=0.15$ \\ 
 & $\text{Value ranking start step}=5k$ & $\text{Value ranking start step}=5k$ & $\text{Value ranking start step}=10k$ \\
 & $\text{Value re-rank size}=128$ & $\text{Value re-rank size}=128$ & $\text{Value re-rank size}=128$ \\ 
\end{tabular}}
\hfill{}
\caption{ 
Hyper-parameter settings. 
}
\label{tab:hyper}
\end{center}
\end{table*}

%% file: 12_path.tex
\section{Value Network Analysis}\label{sec:vna}
We analyzed the ability of the value network to predict expected reward. The reward of a state depends on two properties,  (a) \emph{connectivity}: whether there is a trajectory from this state to a correct terminal state, and (b) \emph{model likelihood}: the probability the model assigns to those trajectories. 
We collected a random set of 120 states in the \scenedomain{} domain from, where the real expected reward was very high ($>0.7$), or very low $(=0.0)$ and the value network predicted well (less than $0.2$ deviation) or poorly (more than $0.5$ deviation). For ease of analysis we only look at states from the final utterance.

%are trajectories that connect it to correct terminal states, (b) model's likelihood: the model's probability of those trajectories. We examine how the value network manages to predict both of those properties and highlight some of its limitations. To this end we collected a set of 120 value network's examples in the \scenedomain{} domain. We considered clear-cut cases where the states relate to the last utterance, their real expected rewards (under the beam search) is either larger than $0.7$ or equals to $0$, and the value network's succeeds to predict the expected reward with high accuracy (less than $0.2$ deviation) or completely fails (more than $0.5$ deviation).

To analyze connectivity, we looked at states that cannot reach a correct terminal state with a single action (since states in the last utterance can perform one action only, the expected reward is 0). Those are states where either their current and target world differ in too many ways, or the stack content is not relevant to the differences between the worlds. We find that when there are many differences between the current and target world, the value network correctly estimates low expected reward in $87.0\%$ of the cases.
%, i.e, both the stack and current world doesn't match the target, or when there are more that 2 mismatches between the current and target worlds. 
However, when there is just one mismatch between the current and target world, the value network tends to ignore it and erroneously predicts high reward in 78.9\% of the cases.    

%We find that the value network 
%correctly estimates if there is a single action (there is only one action left in the last utterance) that will lead the current world to the target world. When there is a single action that leads the current world to the target world, the value network predicts positive reward in 78.9\% of the cases. When multiple actions are necessary, the value function predicts zero reward in 87.0\% of the cases.%

%First, in order to isolate the effect of the connectivity property on the value network estimations, we only take into account states that are not connected to correct terminal states (those states' expected rewards are $0$ regardless of the model's distribution). For states that relate to the last utterance, the connectivity of a state can be determined by checking if the target world can be reached from the current world by applying a single action, and whether the stack holds information that is relevant to this action. We found that the value network manages to predict connectivity only for state where there are multiple discrepancies between them and target world: in $87.0\%$ of those cases the value network output an accurate evaluation, while in $78.9\%$ of the cases where the states  and target world have a single mismatch, the value network makes an inaccurate evaluation.  
To analyze whether the value network can predict the success of the trained policy, we consider states from which there is an action that leads to the target world. While it is challenging to fully interpret the value network, we notice that the network predicts a value that is $>0.5$ in 86.1\% of the cases where the number of people in the world is no more than 2, and a value that is $<0.5$ in 82.1\% of the cases where the number of people in the world is more than 2. This indicates that the value network believes more complex worlds, involving many people, are harder for the policy. 

%Second, we consider the states that are connected to correct terminal states in order to examine the value network's ability to predict the model's likelihood. We found that in $86.1\%$ of the cases where the current world and target world state contain less than 3 people, the value network output is larger than $0.5$, while in $82.1\%$ of the cases where either the current world or target world contain more than 2 people, the value network output is less than $0.5$. The number of elements in the world may indicate in many cases the example's complexity, and therefore the model's probability to parse it successfully. However, this feature may be misleading: in $76.6\%$ of the cases where the value network fails to evaluate states with high value are states with worlds that contain more than 2 people.    

%\section{Programs' Extraction}\label{sec:paths}

%We denote $G$ to be the search-graph found in the execution-search. The paths in $G$ that lead to terminal states represent the discovered programs. A program's correctness is determined by the correctness of its terminal state, i.e whether the state's world matches the target world. 

%We extract correct and incorrect programs separately. Correct programs are built with standard beam search  (guided by our model $\pi_{\theta}$) over prefixes in $G$, that their states are connected to correct terminal states. The search is therefore restricted to the space of the correct found programs. Incorrect programs are built by selecting, for each incorrect terminal state in $G$, a single path that leads to it. 